\documentclass[nohyperref]{article}

\usepackage{microtype}
\usepackage{graphicx}
\usepackage{subfigure}
\usepackage{booktabs} % for professional tables

\usepackage{hyperref}

\usepackage[accepted]{icml2022}

\usepackage{amsmath}
\usepackage{amssymb}
\usepackage{mathtools}
\usepackage{array}
\usepackage{amsthm}

\usepackage[capitalize,noabbrev]{cleveref}

\usepackage[textsize=tiny]{todonotes}

\icmltitlerunning{Solving L2R Challenge by Planning in Latent Space}

\begin{document}

\twocolumn[
\icmltitle{Solving Learn-to-Race Autonomous Racing Challenge\\by Planning in Latent Space}

\icmlsetsymbol{equal}{*}

\begin{icmlauthorlist}
\icmlauthor{Shivansh Beohar}{IIITA} % {equal,IIITA}
\icmlauthor{Fabian Heinrich}{UNIB}
\icmlauthor{Rahul Kala}{IIITA}
\icmlauthor{Helge Ritter}{UNIB}
\icmlauthor{Andrew Melnik}{UNIB}

\end{icmlauthorlist}

\icmlaffiliation{IIITA}{Department of IT, IIIT Allahabad, India}
\icmlaffiliation{UNIB}{Bielefeld University, Germany}

\icmlcorrespondingauthor{Shivansh Beohar}{shivansh.bhr@gmail.com}

\icmlkeywords{Machine Learning, Reinforcement Learning, Planning, Learn2Race,ICML, Safe driving}

\vskip 0.3in
]

\printAffiliationsAndNotice{} % otherwise use the standard text.

\begin{abstract}
Learn-to-Race Autonomous Racing Virtual Challenge hosted on \textit{www.aicrowd.com} platform consisted of two tracks: Single and Multi Camera. Our \textit{UniTeam} team was among the final winners in the Single Camera track. The agent is required to pass the previously unknown F1-style track in the minimum time with the least amount of off-road driving violations. In our approach, we used the \textit{U-Net} architecture for road segmentation, variational autocoder for encoding a road binary mask, and a nearest-neighbor search strategy that selects the best action for a given state. Our agent achieved an average speed of 105 km/h on stage 1 (known track) and 73 km/h on stage 2 (unknown track) without any off-road driving violations. Here we present our solution and results.
\end{abstract}

\section{Introduction}

% Autonomous driving is successfully being deployed in real life, with notable work being done by Tesla, Waymo, Cruise etc in developing consumer ready products. These efforts however, are focused on driving in urban areas and do not completely cater to the skills required in a racing environment, like F1 racing. Recently there has been a surge in research and experimentation in autonomous racing, a similar but new direction to autonomous driving. Due to high costs of testing/deploying hardware and the dangerous environments these products will be deployed in, the safety of the hardware, testing environment and human entities plays a huge role in autonomous racing.
The Learn-to-Race Autonomous Racing Virtual Challenge \cite{herman2021learn}\cite{L2R}\cite{beohar2022planning} provides an F1-style racing environment for a single player to evaluate the performance of self-learning algorithms. An \textit{Open-AI} gym \cite{1606.01540} compliant environment and a \textit{Unreal Engine} based simulator software is provided to develop and test AI agents. In this paper we present our solution and results \cite{L2RWinners}\cite{L2RSolution}. The code implementation is available here: \href{https://gitlab.aicrowd.com/shivansh_beohar/l2r}{\textit{https://gitlab.aicrowd.com/shivansh\_beohar/l2r}}

\section{Challenge and Environment details}

The goal of the challenge is to create autonomous agents which can drive around different tracks in minimum time and with least amount of safety infractions. Safety infractions are described as events in which the agent moves out of the drive-able area, collides with objects like walls or gets stuck somewhere in the track. The primary objective is to get a high success rate, defined as completion \%. The secondary objective is to minimize time taken to complete the track.

The challenge presents a single player racing environment, with several tracks (e.g. \textit{Thruxton, Anglesey National, Las Vegas}) with high fidelity graphics and physics. The observations are multi-modal consisting of $384$x$512$ RGB FPVs of front (Single Camera track), and additional left, right areas (Multi Camera track) of the car plus velocity information. The input action space is steering (right - left) and acceleration (braking - acceleration), both scaled to $[-1,1]$. In our approach, the agent was designed to control the speed rather than the acceleration. The required acceleration was calculated using the current speed and desired speed. 

In the training phase, apart from the RGB views, the simulator provides segmentation views of the cameras with masks for road, car’s hood, ground surrounding the road, and sky. However, the environment did not provide segmentation masks during the evaluation phase.
The competition was split into two stages: 

\begin{itemize}
    \item Stage 1 : Participants have access to the evaluation track (\textit{Thruxton}) and can submit pre-trained models for evaluation.
    \item Stage 2 : Participants only have access to two tracks (\textit{Thruxton} and \textit{Anglesey National}) for developing the local solution, but the evaluation took place on an unknown track (\textit{Las Vegas}). The participants can upload pre-trained models which can be fine-tuned on the unknown track for 1 hour on an \textit{NVIDIA-T4} based compute VM. The fine-tuned model was then evaluated on the unknown \textit{Las Vegas} track and the final metrics was reported on the leaderboard. Participants cannot access the fine-tuned model weights.
\end{itemize}

\begin{figure}[htbp]
\centerline{\includegraphics[scale=0.7]{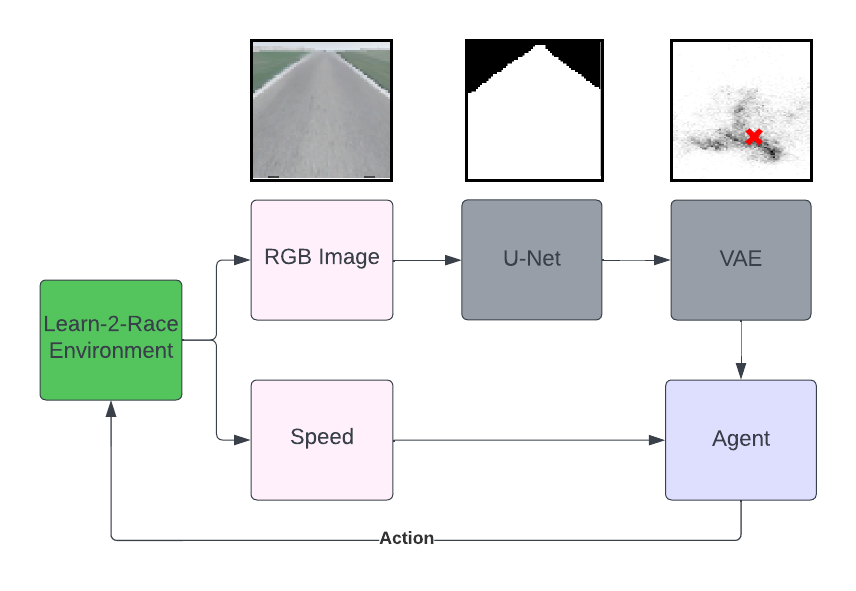}} 
%TODO Learn-2-Race --> Learn-to-Race; bold font; Action - the same font size as other text; crop white boarders
%TODO add RGB --> Critic --> VAE stream and add value colors to the 2D-latent space
\caption{The data-flow pipeline. The pre-processed and resized RGB image arrives at the \textit{U-Net}. The \textit{U-Net} output is a binary mask of the road, reduced to $28$x$28$ and passed to \textit{VAE}. \textit{VAE} performs a two-dimensional latent encoding, red cross represents the encoding of the present binary mask. This encoding, along with speed information from the environment, is used by the agent as a state representation to produce an action.}
\label{pipeline}
\end{figure}

%We then derived a process which identifies the steering actions to take corresponding to different turns, which are the state encoding of visuals where the road takes a turn. The process is described in the following subsections.

\section{Scene Semantics and State Encoding}
Use of semantics as a part of input is a widely adopted approach in autonomous control tasks. Semantics are the important entities in the scene represented often as segmentation masks or coordinates. Using semantic masks instead of (or in combination with) RGB images helps to reduce state complexity while keeping the necessary information in the state representation. Using VAE \cite{kingma2013auto} to encode the semantics further help in creating a smooth latent space, which enables better interpolation between the state representations, leading to better noise agnostic learning.  \cite{9263921} demonstrate a similar End-to-End architecture to train DDPG \cite{article} on CARLA town-I task. However, deep neural network based learning algorithms including \cite{9263921} require massive amount of data and time to learn good behaviors. They result in robust and high performance, but at the cost of training time and poor interpretability. Fast learning algorithms were a prerequisite for the \textit{L2R} challenge. We achieved fast learning by using semantically sufficient state encoding and searching for safe behaviors in the replay buffer. See section \ref{Discussion} for limitations of our approach.

\section{Methods}

\subsection{U-Net}
We trained the \textit{U-Net} model \cite{ronneberger2015u} to convert RGB images to road masks during the training phase on segmentation masks of car hood, sky, grass, and road provided by the environment. To make the \textit{U-Net} model generalize on unknown tracks, we used color-based image augmentation to augment the collected images using the \textit{Albumentations} library \cite{info11020125}. Specifically we used \textit{ColorJitter (p=0.2)}, \textit{Hue(p=0.2)}, \textit{Saturation(p=0.2)}, \textit{Contrast(p=0.2)}, \textit{RGB shift(p=0.5)}, \textit{Channel Shuffle(p=0.5)}, \textit{CLAHE (p=0.2)}, and \textit{Sepia(p=0.2)} for color transformations along with random horizontal flipping, scaling, shifting, and rotating (p=0.5) for view transformations, where p indicates the probability of applying the transformation.

\subsection{VAE}
We trained a variational auto-encoder (\textit{VAE}) \cite{kingma2013auto} to encode the road-mask images into two-dimentional latent encodings. The dataset for training of \textit{VAE} was collected from runs in the environment in which the agent drives the car at constant speed in a straight line.

\subsection{State representation pipeline}
The data-flow pipeline is depicted in Fig. \ref{pipeline}. The 384x512 RGB images provided by the \textit{L2R} environment are cropped (lower and upper stripes) to remove the hood and the sky, producing a $100$x$512$ image which contains the essential information for steering. This is resized to $64$x$64$ and fed into the pre-trained \textit{U-Net} model. The output of the \textit{U-Net} model is a binary mask $64$x$64$ of the road. The mask is resized to $28$x$28$ and fed to the \textit{VAE} to produce a 2-dimensional latent encoding. This encoding, along with the speed information from the environment, is used by the agent as the state representation for producing the action output.

\subsection{Environment's reward function}
The environment returns positive reward value for every time step proportional to the speed of the agent and negative reward value for driving off the road:\\min(-1*25, -1*5*velocity)

\subsection{Base behavioural policy and value function discriminator} 

\begin{figure}[htbp]
\centerline{\includegraphics[scale=0.2]{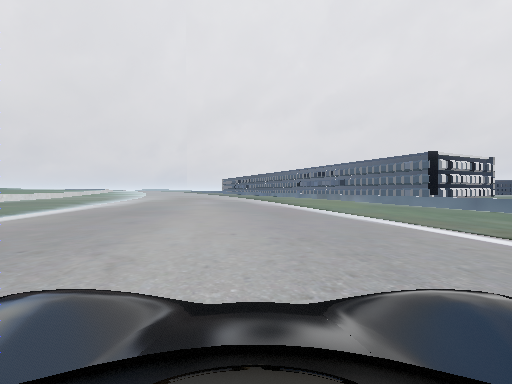}
\includegraphics[scale=0.2]{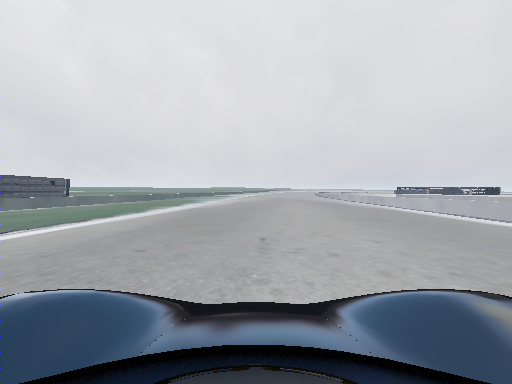}}
\centerline{\includegraphics[scale=0.2]{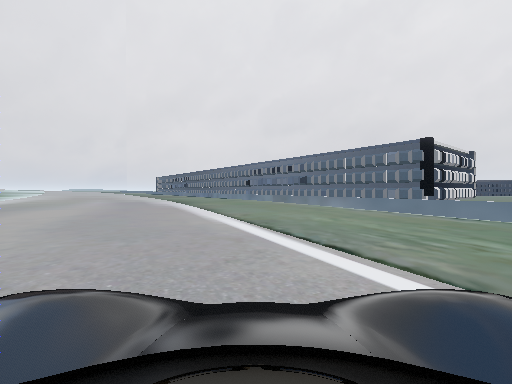}
\includegraphics[scale=0.2]{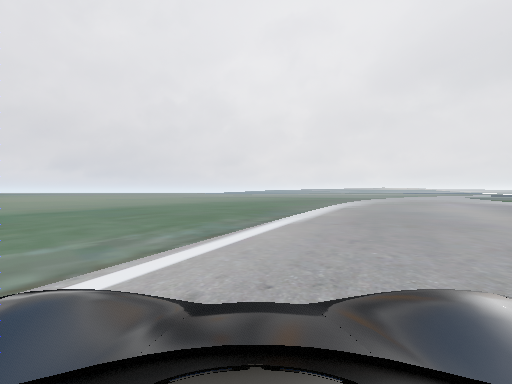}}
\caption{The first row - \textit{safe} state examples. The second row - \textit{unsafe} state examples.}
\label{unsafe_safe_eg}
\end{figure}

The most frequently used action from the demonstration dataset, namely the \textit{drive straight} action, served as the agent's base behavioral policy. A dataset of training trajectories was collected using this base behavioral policy (\textit{drive straight}) for 20 episodes, obtaining  of about 1200 states altogether. Depending on the initial heading angle and the distance to the edge at the beginning of each episode (fig. \ref{unsafe_safe_eg}), this results in the agent driving along the road for some time steps and driving off the road at the end of each episode, thus completing the episode. Using the collected reward, the agent computed the discounted accumulated reward estimate for each state in the trajectories.

The value function is implemented as a five-nearest-neighbors classifier. If the average value of five-nearest-neighbors is positive, then the current state is \textit{safe} and the agent continues to follow the base behavioural policy, \textit{drive straight}. If the average value is negative, then the current state is \textit{unsafe}, and the agent follows the correction policy (see section \ref{cpolicy}).

We also experimented with training a convolution neural network for predicting the value estimation of RGB-image input, assuming the base behavioural policy the agent drives straight. This neural network model produced similar results.

\subsection{Correction policy: action discriminator}
\label{cpolicy}

Training of the correction policy occurs through execution of random discrete actions whenever a state has become \textit{unsafe}. A randomly selected action is executed until the state becomes \textit{safe} or until the episode ends. If the selected action executed for some number of time steps leads to a \textit{safe} state, these state-action pairs are labeled as \textit{good} and saved in the correction policy buffer, if lead to the end of episode, then these state-action pairs are discarded. The correction policy buffer was trained for another 20 episodes.

Exploitation of the correction policy works though the five-nearest-neighbors search in the correction policy replay buffer of \textit{good} transitions. The discrete action is selected by the majority of five-nearest-neighbors in the current \textit{unsafe} state.

The correction policy is selected to control the agent until the value function discriminator classifies the current state as \textit{unsafe}. After the current state becomes \textit{safe} again, the agent selects the base behavioural policy to control the agent.

\subsection{Evaluation}
During the evaluation phase, the agent selects the base behavioural policy to control the agent in \textit{safe} states and the correction policy in \textit{unsafe} states. The value function determines whether the current state is \textit{safe} or \textit{unsafe}.

% We set two levels of speed, the higher being 90 km/h which is to be used in safe states and the lower being 60 km/h which is to be used in unsafe states.

% With the obtained policy, the agent could complete any track (seen/unknown) at around 62 km/h in the unknown track and around 78 km/h in seen tracks. The difference is probably due to the difference in the tracks, the seen track offers continuous high speed segments with some choke points, while the unknown track offers continuous hard and soft turns, leading to a lot of braking.

% \section{Improvements}
% We made the following two improvements to the agent, in order to better switch between different velocities while running on a track.

\subsection{Speed adaptation}
%the agent learned to segment the road mask from RGB input images using the U-Net architecture and 

In the one hour given to the agent to adapt to the competition track, the agent gradually increased its speed in episodes to find the maximum speed level at which it would not go off the road. It searched for the optimal speed in \textit{safe} and \textit{unsafe} states. The agent had an internal model of the distance from the beginning of the episode to the current time step, which can be approximated by the number of time steps and the speed at each time step. In this internal model, the speed for positive and negative episodes were accumulated and then an upper bound on the \textit{safe} speed for a given distance segment from the starting position of the track was estimated. In training runs, the agent increased speed for each distance segment it had survived in the previous run and decreased it for distance segments where it did not survive in previous runs. We found 15 runs to be enough for the purpose of speed adaptation.

%\subsection{Exponential Braking}
% On the straight segments of track, unsafe states are rare. This causes a speed reduction to 60 km/h, that is almost immediately undone by the next safe state. 
%This has shown to cause a jerky motion on smooth curved sections of the track, leading to decreased average speed. 

%We increased the maximum speed achievable for the agent to be 126 kmh, so that it can benefit from better braking.
%With this improvement, the performance increased to about 98 km/h on the known tracks and 66 km/h on the unknown track. 

\begin{table}

\begin{tabular}{ | m{4.0cm} | m{3.0cm} | }
\toprule
Track  & Avg. speed (km/h)\\
\midrule
Thruxton (known)       & \textbf{112.43 $\pm$ 8.8}       \\
\hline
Anglesey National (known)    & \textbf{90.34 $\pm$ 8.3}  \\
\hline
Las Vegas (unknown) & \textbf{73.19} \\ 
\bottomrule
\end{tabular}
\caption{Performance comparison on different tracks \\ \* Only one data point is available for \textit{Las Vegas} track since we do not have access to the local version of the track. All other readings are an average of 5 evaluations, with 3 laps each.}
\label{tab:result}
\end{table}

\section{Results}

While testing, the agent maintained the maximum \textit{safe} speed for a given distance segment according to the learned internal model. Thus the agent achieved the average speed of 73 km/h on the unknown track and to 112 km/h on the known track.

The average speed achieved is shown in Table \ref{tab:result}. Due to limited access to \textit{Las Vegas} track, only leaderboard scores \cite{L2R} are reported. All the experiments achieved 100\% success rate in all the tracks.

% \section{Other Experiments}
% In the course of the competition, we tried out various other RL techniques.
%which proved to helpful, but were not used in the final solution due to some difficulties, as highlighted in the subsections below.

% -- % -- 

\section{Discussion}
\label{Discussion}
In this paper we present our solution which matches latent space representation of states to control the car and optimizes the mean speed. One drawback of our solution is that it is only able to predict on a short horizon, which does not allow human-like performance by choosing the optimal curved path on turns. Another drawback of out solution is that it works well for low dimensional noise-free state representations, where KNN lookup is feasible. Alternatively, control tasks with rich sensory information can be solved with deep reinforcement learning techniques \cite{melnik2021critic}\cite{bach2020error} by learning a policy \cite{konen2019biologically}\cite{schilling2018approach}\cite{schilling2021decentralized}. In the course of the competition, we tried out various other RL techniques \cite{bach2020learn} like SAC, DQN etc., but they turned out either very expensive to train or resulted in low average speeds due to zig-zag driving. There is a potential to extend our solution by using Monte Carlo Tree Search to evaluate different paths \cite{harter2020solving}, and modularization of the task for efficient learning \cite{melnik2019modularization}, while maintaining safety constraints.

We would like to thank the organizing team \cite{herman2021learn} for the technical support provided throughout the competition.

\bibliography{refs}
\bibliographystyle{icml2022}
\end{document}